\documentclass{article}
\usepackage{spconf,amsmath,graphicx}
\usepackage{booktabs}
\usepackage[ruled,vlined]{algorithm2e}
\usepackage{svg}
\usepackage{hyperref}
\hypersetup{
    colorlinks=true,
    linkcolor=blue,
    filecolor=blue,      
    urlcolor=blue,
}
\usepackage{enumitem}
\setlist{nosep, leftmargin=14pt}

\usepackage{mwe} 
\usepackage{xcolor}

\DeclareMathOperator*{\argmax}{arg\,max}

\title{Point-supervised Segmentation of Microscopy Images and Volumes via Objectness Regularization}
\name{%
\begin{tabular}{@{}c@{}}
Shijie Li$^{\star}$\thanks{Corresponding author: shijie.li@nyu.edu.} \qquad 
Neel Dey$^{\star}$ \qquad 
Katharina Bermond$^{\ddagger}$\\ 
Leon von der Emde$^{\dagger}$ \qquad 
Christine A. Curcio$^{\dagger\dagger}$ \qquad 
Thomas Ach$^{\dagger}$ \qquad
Guido Gerig$^{\star}$
\end{tabular}}

\address{$^{\star}$ Computer Science and Engineering, New York University, Brooklyn, NY, USA. \\
$^{\ddagger}$Ophthalmology, Ludwigshafen Hospital, Ludwigshafen, Germany\\
$^{\dagger}$ Ophthalmology, University Hospital Bonn, Bonn, Germany\\
$^{\dagger\dagger}$Ophthalmology and Visual Sciences, University of Alabama at Birmingham, Birmingham, AL, USA}
\begin{document}
%
\maketitle
\begin{abstract}

Annotation is a major hurdle in the semantic segmentation of microscopy images and volumes due to its prerequisite expertise and effort. This work enables the training of semantic segmentation networks on images with only a single point for training per instance, an extreme case of weak supervision which drastically reduces the burden of annotation. Our approach has two key aspects: (1) we construct a graph-theoretic soft-segmentation using individual seeds to be used within a regularizer during training and (2) we use an objective function that enables learning from the constructed soft-labels. We achieve competitive results against the state-of-the-art in point-supervised semantic segmentation on challenging datasets in digital pathology. Finally, we scale our methodology to point-supervised segmentation in 3D fluorescence microscopy volumes, obviating the need for arduous manual volumetric delineation. Our code is freely available.

\end{abstract}
\section{Introduction}
\label{sec:intro}

Real-world analysis of microscopy images and volumes involves the manual review of terapixels of content by trained experts. Cellular and nuclear morphology obtained via segmentation are phenotypic reflections of underlying genotypic variation and inform relevant decisions in disease treatment and drug design. Currently, deep networks trained on datasets of expert annotations are widely deployed towards accelerating research throughput. However, the construction of training sets with the scale required for deep learning-based image segmentation is an often insurmountable investment of time and resources for some study designs. These hurdles increase exponentially for volumetric images which are much harder to annotate, especially in modalities with densely-clustered instances of interest such as lysosomes and related organelles.

Practically, researchers employ one or both of the following strategies: (1) transfer learning via pre-training on large public collections of labeled images, or (2) employing weak-supervision in training. While established, transfer learning in microscopic segmentation often faces questions of generalizability as appearance is modality-sensitive. Furthermore, publicly available fully annotated volumetric microscopy images for semantic segmentation are exceedingly rare. Thus, we focus on the weak supervision perspective in this work.

Various strategies have been developed for semantic segmentation from local and/or global weak annotation, such as scribbles \cite{lin2016scribblesup}, bounding-boxes \cite{khoreva2017simple}, corner points \cite{maninis2018deep}, among others. If scribble annotation is eroded to a single mark on an object of interest, we arrive at a point-supervised segmentation setting. Point-supervision greatly alleviates the burden of annotation and has therefore received significant attention.

Relevant to our work, \cite{bearman2016s} creates foreground saliency maps using a pre-trained network and trains segmentation with a partial cross-entropy objective and a saliency-based regularizer. \cite{tang2018regularized,tang2018normalized} find that relaxations of standard affinity-based regularizers benefit training with partial cross-entropy. \cite{laradji2018blobs} proposes a hybrid loss function that predicts blobs for point-annotated instances while suppressing false positives. \cite{pmlr-v102-qu19a} constructs a heuristic training set with a combination of Voronoi triangulation and K-means clustering to train a segmentation network with conditional random field refinement. Most recently, \cite{yoo2019pseudoedgenet} uses auxiliary networks to learn edge information via a Sobel-filter based loss. Notably, these methods have only been demonstrated in planar segmentation and have not yet been shown to scale to 3D.

In this work, we present a method for point-supervised semantic segmentation leveraging key insights in microscopic image analysis - objects of interest are often convex and are weakly homogeneous in intensity. We construct an image graph using custom affinity functions and use iterated maximum flow \cite{schrijver2002history} from individual seeds to create a soft-segmentation prior of the image to use as a regularizer. We train on a combination of this \textit{objectness} and the labeled points using a custom objective function. Using our framework, competitive results against the state-of-the-art are achieved on a variety of digital pathology datasets. Finally, we scale our method to 3D images and greatly outperform models trained with conventional masked losses. Our PyTorch code is available \href{https://github.com/CJLee94/Point-Supervised-Segmentation}{here}.

\section{Methods}

\begin{figure}[t]
    \centering
    \includegraphics[width=0.45\textwidth]{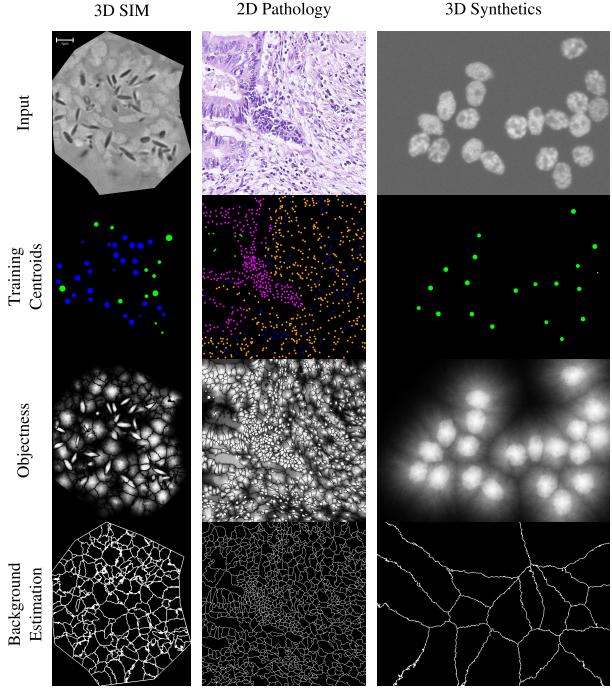}
    \caption{Given a training set of input images (\textbf{row 1}) and locations of target object centroids (\textbf{row 2}), we generate soft probabilistic multi-class \textit{objectness} maps (\textbf{row 3}) used to regularize segmentation training in the weakly supervised regime. The boundaries of image sub-regions estimated by the objectness algorithm are treated as background labels (\textbf{row 4}).}
    \label{fig:objectness_samples}
\end{figure}

Here we motivate and detail methodology towards training general segmentation networks via a multi-class objectness used for regularization generated prior to training and a multi-task loss function which enables weakly supervised training. \\

\begin{figure}[t]
    \centering
    \includegraphics[width=0.45\textwidth]{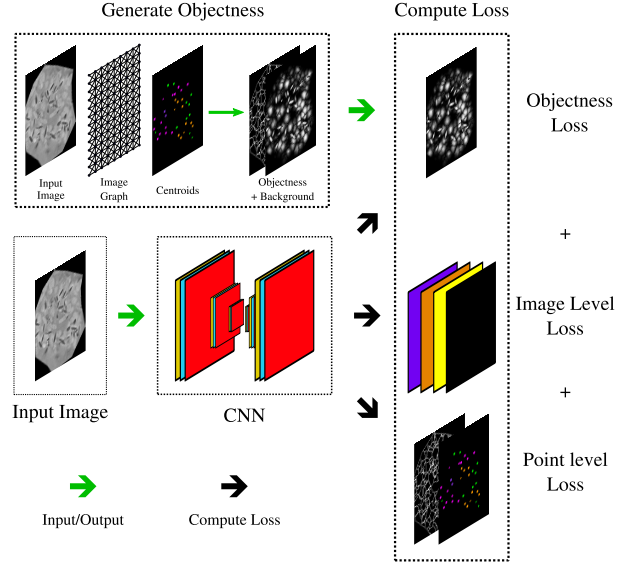}
    \caption{Overall pipeline of our work}
    \label{fig:my_label}
\end{figure}



\begin{figure*}[!ht]
    \centering
    \includegraphics[width=0.9\textwidth]{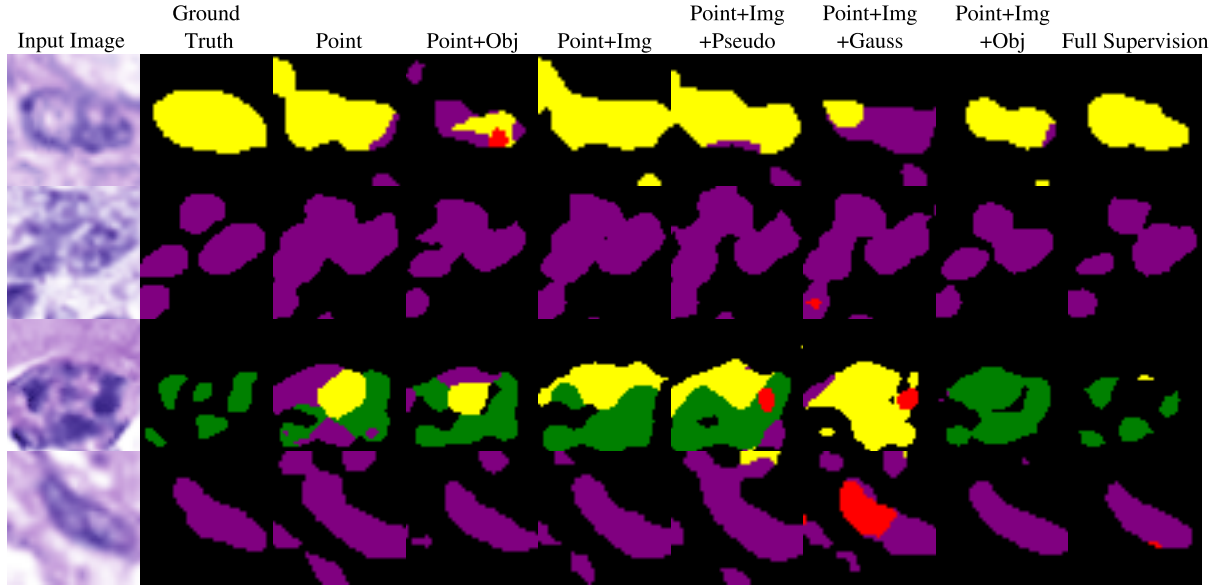}
    \caption{Results of point-supervised multi-class segmentation for CoNSeP dataset.}
    \label{fig:2d_seg_results}
\end{figure*}

\noindent\textbf{Objectness.} \textit{Objectness} (also known as \textit{saliency}) is defined here as the likelihood of whether a pixel belongs to the foreground. In point-annotated natural image segmentation, \cite{xiong2018pixel,bearman2016s}  obtain objectness maps via ImageNet-pretrained networks. As no such pretrained networks exist for  nuclei/organelle segmentation, we propose a novel methodology to generate objectness maps from multi-class point annotation. Intuitively, pixels spatially and chromatically close to a point annotation marking an object are more likely to be part of the object compared to pixels further away from the annotation. We propose an objectness generation method loosely inspired by the Watershed algorithm, as below:

\begin{enumerate}
    \item Give each annotated seed point a unique identifier such that pixels in a region grown from the seed (referred to as \textit{identified} pixels) share its identifier.
    \item Push the currently identified pixels (initially only the seeds) into a priority queue primarily in order of their distance to the seed pixel and secondarily in order of the time of entering the queue. Here, \textit{distance} reflects spatial and chromatic dissimilarity and is defined recursively as,
    \begin{align}
        \label{eq:distance}
        \begin{split}
            d(seed(p), p) &= d(seed(p), par(p))+d_I(p, par(p)),\\
            d_I(p) &= \|I_p-I_{par(p)}\|_2^2,
        \end{split}
    \end{align}
    where $d$ is the cumulative distance from the seed pixel $seed(p)$ that the region grows from to the pixel $p$ and $I_p$ refers to pixel intensities. $par(p)$ is defined below. 
    \item Pop the identified pixel with the shortest distance to a seed point (if tied, use the earlier time of queue entry). Assign neighboring pixels the same identifier as the popped pixel and push them into the priority queue. The popped pixel is the parent ($par(p)$) of its neighboring pixels.
    \item Repeat step 3 until the queue is empty.
\end{enumerate}
Following the above, the image is divided into distinct regions with different identifiers. Pixels that lie on region boundaries are now treated as background annotations. Once each pixel has an associated distance map, the objectness is calculated as $Obj(p) = e^{-w d(seed(p), p)}$ where $w$ is a hyperparameter. As opposed to the Watershed algorithm, which assigns hard labels, the proposed approach generates soft-labels allowing for uncertainty. A visual overview of objectness inputs and outputs across datasets is given in Figure \ref{fig:objectness_samples}.

\noindent\textbf{Loss Function.} We use multiple losses incorporating various forms of weak supervision and use the same notation as \cite{bearman2016s} for consistency. Let $S_{ic}$ be the probability of the $c$-th class at the $i$-th pixel and the output be of dimension $N\times C\times D_1, D_2, \dotsc$, where $N$ is the batch size, $C$ is the number of classes, and $D_1, D_2,\dotsc$ are the image dimensions.

\noindent\textit{\underline{Point-Level Supervision.}} 
Given point annotations, cross-entropy can be used to train multiclass segmentation by masking all unlabeled pixels in the loss. Sparse background annotation for negative sampling is generated for all methods using the methods described in the above objectness section. To avoid overfitting to the generated background labels, we downweigh the background samples (0.1). Here $G_i$ is the class that pixel $i$ belongs to in the ground truth label:

\begin{equation}
    \mathcal{L}_{\text{point}}(S,G,F,B) = -\sum_{i\in \mathcal{I}_s}\alpha_i\log(S_{iG_i}).
\end{equation}


\noindent\textit{\underline{Objectness Supervision.}}
Given the generated probabilistic multi-class objectness, we compute the weighted softmax cross-entropy of the objectness against the network prediction as in \cite{bearman2016s}. We use cross-entropy as opposed to a soft-Dice loss as cross-entropy is relatively robust to label noise \cite{rolnick2017deep}: 
\begin{equation}
    L_{\text{obj}}(S,P) = -\frac{1}{|I|}\sum_{i\in I}\left(\sum_{c\in C}\beta_cP_{ic}\log(S_{ic})\right),
\end{equation}
where $P_{ic}$ is the objectness of pixel $i$ for class $c$. We add $\beta_c$ here to avoid over-segmentation by setting $\beta$ for the background class to be higher than other classes.

\noindent\textit{\underline{Image-Level Supervision.}}
As is common in weakly supervised multi-class segmentation works\cite{bearman2016s}, our networks are further trained to predict whether an object class is present in the image or not. Let $L$ be the set of classes that present in the image, while $L'$ be those not present in the image. The loss function becomes as follows, where $t_c = \argmax_{i\in\mathcal{I}} S_{ic}$,

\begin{equation}
    \mathcal{L}_{\text{img}}(S,L,L') = -\frac{1}{|L|}\sum_{c\in L}\log S_{t_cc}-\frac{1}{|L'|}\sum_{c\in L'}\log(1-S_{t_cc}).
\label{eq:imglvl_loss}
\end{equation}
The overall objective function is $\mathcal{L} = \lambda_1\mathcal{L}_{\text{point}}+\lambda_2\mathcal{L}_{\text{obj}}+\lambda_3\mathcal{L}_{\text{img}}$, where $\lambda_1 = \lambda_2 = \lambda_3 = 1$ in our experiments.


\section{Experiments}

\noindent\textbf{Datasets.} For benchmarking the proposed methodology, we use a diversity of public datasets across modalities. In 2D digital pathology, we use the colorectal nuclear segmentation and phenotypes (\textit{CoNSeP}) dataset \cite{graham2019hover} containing 41 $1000 \times 1000$ hematoxylin and eosin stained (H\&E) images at 40X magnification.
As there exists no publicly available large-scale 3D microscopy dataset with full annotation to our knowledge, we quantify using the BBBC024 synthetic fluorescence microscopy dataset from the Broad Bioimage Benchmark Collection of HL60 cell nuclei
\cite{ljosa2012annotated}.

Our driving biomedical application focuses on the delineation of organelle morphology in the human retinal pigment epithelium (RPE). Human donor RPE flatmounts were imaged with super-resolution structured illumination microscopy (SIM) on a Zeiss Elyra S.1 (488 nm excitation) as part of a recent ophthalmological study \cite{bermond2020autofluorescent} upon which we apply our approach.
Morphological variation (size, shape, localization, density, and distribution) of RPE granules such as lipofuscin, melanolipofuscin, and melanosomes are associated with normal aging and disease progression \cite{ach2014quantitative,ach2015lipofuscin} and can be tracked via automated segmentation.
As each RPE cell contains several hundred organelles, full 3D annotation of large numbers of RPE cells per tissue sample is not feasible. Our biomedical partners localized and classified each organelle per cell with point annotations.

\noindent\textbf{Implementation details.} Dataset-specific design choices and hyperparameters are given in the relevant sections. We use Adam for optimization throughout with a step decay when validation loss plateaus. Bayesian Optimization
is used to find optimal parameters in terms of IoU for objectness generation on the validation sets. We do not pretrain networks.

\noindent\textbf{Baselines.} For 2D benchmarks, we compare against ablations of our full model and a recently proposed point-supervised method (\textit{Pseudo}) \cite{yoo2019pseudoedgenet}. An objectness based on 2D Gaussian representations was proposed in \cite{hofener2018deep} for nuclei detection (\textit{Gauss Obj}) and is included here to compare against our objectness generation. We are the first to benchmark 3D point-supervised segmentation to our knowledge and thus only compare against using partial cross-entropy (\textit{pCE}) alone.  \\

\begin{figure}[t]
    \centering
    \includegraphics[width=0.48\textwidth]{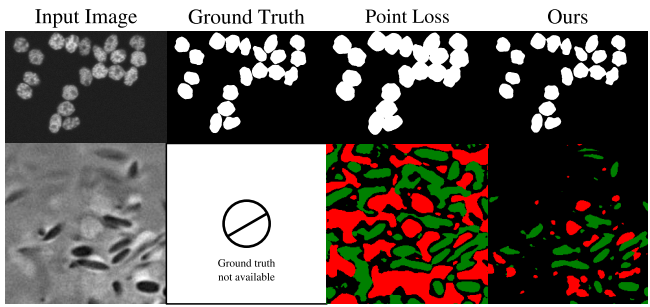}
    \caption{Selected samples of segmentation results on the synthetic BBBC dataset and the ex-vivo human SIM dataset.}
    \label{fig:3dresult}
\end{figure}

\noindent\textbf{2D Digital Pathology Nuclei Segmentation.} For CoNSeP, we perform a 27/7/7 split at the image-level. Consistent with \cite{yoo2019pseudoedgenet}, we use random affine transformations, HSV-space shifts, blurring, and Gaussian noise injection for data augmentation. During training, we randomly crop patches of size $128\times 128$ using a batchsize of $64$. The objectness algorithm relies on the assumption that target pixel intensity is relatively homogeneous, thus we select an arbitrary target image and normalize image stains  \cite{vahadane2016structure} to the target for objectness generation. For objectness, we perform anisotropic diffusion for edge-preserving smoothing prior to generation. 

\textit{\underline{CNN Model:}} Here, we use Feature Pyramid Networks (FPN) \cite{lin2017feature} which reuse multiscale features for improved segmentation performance. The backbone is a ResNet50 architecture and concatenation is used for decoder merging. The initial learning rate is set to $10^{-3}$ for every method.

\textit{\underline{Results:}}
Quantitative results for point-supervised multi-class segmentation against baselines and ablations are shown in Table \ref{table:2dresults} with segmentation results visualized in Figure \ref{fig:2d_seg_results}. We see that the our weakly supervised method closes the gap between the
fully supervised upper bound and the masked loss only model and compares favorably to recent baselines. \\ 

\begin{table}[t]
\centering
\caption{2D \textit{multi-class} point-supervised nuclei segmentation results in terms of mean IoU (higher is better).}
\begin{tabular}{c@{\hskip 0.1in}c@{\hskip 0.1in}c}
    \toprule
    \textbf{Experiment} & \textbf{Loss function} & \textbf{mIoU} \\
    \midrule
    Ablation & pCE & $0.4596\pm0.0350$ \\
     & pCE+Img & $0.5450\pm0.0344$\\
     & pCE+Obj & $0.4973\pm0.0307$\\ \hline
    Baselines & +Pseudo \cite{yoo2019pseudoedgenet} & $0.5357\pm0.0296$\\
    (pCE+Img) & +Gauss Obj \cite{hofener2018deep} & $0.5491\pm0.0326$ \\ 
    \hline
    Proposed & pCE+Img+Obj & $\mathbf{0.5546\pm0.0312}$ \\ \hline
    Upper Bound & Full Supervision & $0.5837\pm0.0347$ \\
    \bottomrule
\end{tabular}
\label{table:2dresults}
\end{table}

\noindent\textbf{3D Fluorescence Microscopy Segmentation} We split at the image level for BBBC024 with a 20/5/5 train/validation/test split. For SIM, we split at the cell level (with each cell containing hundreds of organelles) arriving at a 199/24/24 split. We use discrete transformations sampled from the $D_{4h}$ group for augmentation. Random crops of $64 \times 64 \times 32$ are used for training, with a batch size of 20. We do not perform any preprocessing on BBBC024 and use only histogram equalization on the SIM dataset. When generating objectness for SIM, we perform anisotropic diffusion smoothing.

\textit{\underline{CNN Model:}} We use a 3D U-Net with 4 levels and two $3\times 3$ Conv/BatchNorm/ReLU blocks with 64 filters per level.

\textit{\underline{Results:}} On BBBC024, we find that a segmentation model trained with partial cross entropy alone achieves an IoU of 0.3238, but when combined with our objectness regularization performance reaches 0.6696, with a fully supervised upper bound of 0.8816. As full volumetric delineations are not available for the SIM dataset, we display qualitative results in Fig.~\ref{fig:3dresult}, finding that our method provides high-quality 3D segmentation training only on point annotations.

\section{Discussion}
We present a network-agnostic segmentation regularizer which when combined with an appropriate loss function enables point-supervised semantic segmentation in diverse planar and volumetric microscopy modalities. Competitive mIOU scores are achieved for point-supervised digital pathology images against recent baselines. Finally, we demonstrate the viability of this approach in 3D segmentation both in synthetic HL60 images and in SIM images of RPE cells, enabling future high throughput volumetric analysis. 

\section{Acknowledgements}
The authors were supported by NIH R01EY027948. TA was further supported by Dr. Werner Jackstädt Foundation and CAC was additionally funded by Heidelberg Engineering and Genentech. TA and CAC are shareholders of MacRegen Inc.

\section{Compliance with Ethical Standards}
Retrospective analysis was performed on two open-access datasets \cite{graham2019hover,ljosa2012annotated} and a database associated with \cite{bermond2020autofluorescent}, all of whom complied with the Declarations of Helsinki and received approval at their respective institutions.




\bibliographystyle{IEEEbib}
\bibliography{refs}

\end{document}